\documentclass[10pt,twocolumn,letterpaper]{article}

\usepackage{wacv}
\makeatletter
\@namedef{ver@everyshi.sty}{}
\makeatother
\usepackage{epsfig}
\usepackage{graphicx}
\usepackage{amsmath}
\usepackage{amssymb}

\usepackage{algorithm}
\usepackage{diagbox}
\usepackage{algorithmicx}
\usepackage{algpseudocode}
\usepackage{mathtools}
\usepackage{amsfonts}
\usepackage{siunitx} 
\usepackage{textcomp}
\usepackage{algpseudocode}

\usepackage{comment}

\usepackage{tabularx}
\usepackage{lipsum}
\usepackage{gensymb}
\usepackage{multirow,enumitem}

\usepackage{tikz}
\usetikzlibrary{shapes,arrows}
\usetikzlibrary{fit}
\usetikzlibrary{matrix}

\usepackage[percent]{overpic}
\usepackage{xcolor}

\wacvfinalcopy 

\def\wacvPaperID{680} 

\ifwacvfinal\fi
\begin{document}

\title{Gaze Estimation for Assisted Living Environments}

\author{Philipe A. Dias$^{1}$ \hspace{0.5cm} Damiano Malafronte$^{2,3}$ \hspace{0.5cm} Henry Medeiros$^{1}$ \hspace{0.5cm}
Francesca Odone$^{2}$ \\
{\small$^{1}$Marquette University (EECE), USA  \hspace{0.5cm} $^{2}$University of Genoa, Italy \hspace{0.5cm}$^{3}$Italian Institute of Technology(IIT)}\\ 
\hspace{-2pt}{\tt\small\{philipe.ambroziodias,henry.medeiros\}@marquette.edu\hspace{0.4cm}damiano.malafronte@iit.it\hspace{0.25cm} francesca.odone@unige.it}
}

\maketitle
\ifwacvfinal\thispagestyle{empty}\fi

\begin{abstract}
Effective assisted living environments must be able to perform inferences on how their occupants interact with one another as well as with surrounding objects. To accomplish this goal using a vision-based automated approach, multiple tasks such as pose estimation, object segmentation and gaze estimation must be addressed. Gaze direction in particular provides some of the strongest indications of how a person interacts with the environment. In this paper, we propose a simple neural network regressor that estimates the gaze direction of individuals in a multi-camera assisted living scenario, relying only on the relative positions of facial keypoints collected from a single pose estimation model. To handle cases of keypoint occlusion, our model exploits a novel confidence gated unit in its input layer. In addition to the gaze direction, our model also outputs an estimation of its own prediction uncertainty. Experimental results on a public benchmark demonstrate that our approach performs on pair with a complex, dataset-specific baseline, while its uncertainty predictions are highly correlated to the actual angular error of corresponding estimations. Finally, experiments on images from a real assisted living environment demonstrate the higher suitability of our model for its final application.
\end{abstract}
\section{Introduction}
The number of people aged 60 years or older is expected to nearly double by 2050 \cite{UN2017World}. The future viability of medical care systems depends upon the adoption of new strategies to minimize the need for costly medical interventions, such as the development of technologies that maximize health status and quality of life in aging populations. Currently, clinicians use evaluation scales that incorporate mobility and Instrumented Activities of Daily Living (IADL) assessments (i.e., a person's ability to use a tool such as a telephone without assistance) \cite{pilotto2008development} to determine the health status of elderly patients and to recommend changes of habits. 

Despite the potential of recent advances in many areas of computer vision, no current technology allows automatic and unobtrusive assessment of mobility and IADL over extended periods of time in long-term care facilities or patients' homes. Patient activity analysis to date has been limited to simplistic scenarios \cite{debes2016monitoring}, which do not cover a wide range of relatively unconstrained and unpredictable situations.

\begin{figure}[h]
	\centering
	\includegraphics[width=\linewidth]{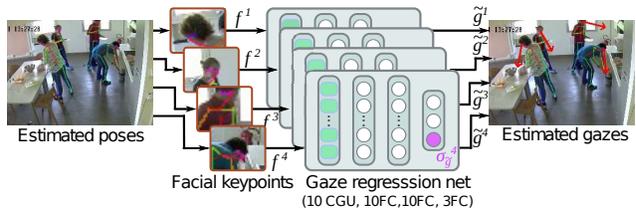} 
	\caption{Overview of our apparent gaze estimation approach. The anatomical keypoints of all the persons present in the scene are detected using a pose estimation model \cite{cao2017realtime}. The facial keypoints of each person are then provided as inputs to a neural network regressor that outputs estimations of their apparent gaze and its confidence on each prediction.}
	\label{fig:method_diagram}
\end{figure}

\begin{figure*}[t]
    \includegraphics[width=\linewidth]{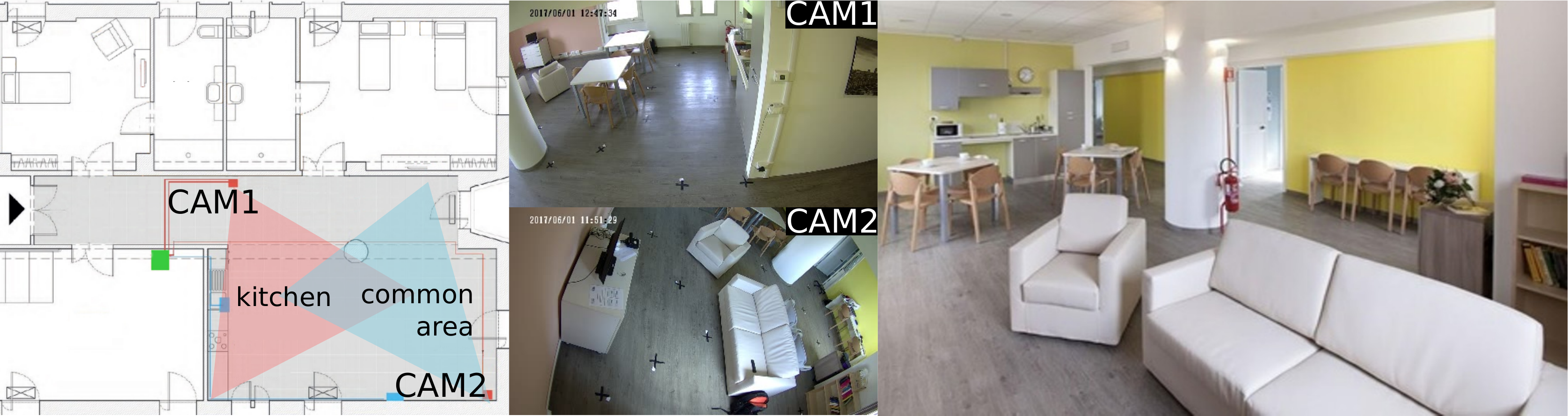}
\caption{Images and layout of the instrumented assisted living facility; in color, the video-cameras fields of views. 
}
\label{fig:apt_plan}
\end{figure*}

Vision-based analysis of mobility and characterization of ADLs is rather challenging. As examples in Fig.~\ref{fig:method_diagram} and \ref{fig:apt_plan} illustrate, images acquired from assisted living environments cover a wide scene where multiple people can be performing different activities in a varied range of scenarios. Moreover, it encompasses multiple underlying complex tasks including: detection of subjects and objects of interest, identification of body joints for pose estimation, and estimation of the gaze of the subjects in the scene.

In this paper we focus on gaze estimation, which is a critical element to determine how humans interact with the surrounding environment. It has been applied to design human-computer interaction methods \cite{majaranta2014eye} and to analyze social interactions among multiple individuals \cite{Varadarajan2018}. For our application, in conjunction with object detection \cite{dias2017fine}, gaze direction could define mutual relationships between objects and their users (e.g. the user is sitting on a chair with a book on his/her lap vs. sitting on a chair reading the book) and classify simple actions (e.g. mopping the floor, getting dressed, reading a book, cooking food, eating/drinking).


The contributions of the present work can be summarized in three main points:
\begin{itemize}[leftmargin=*,topsep=0pt,itemsep=.25mm]
    \item \textit{we propose an approach that relies solely on the relative positions of facial keypoints to estimate gaze direction}. As shown in Fig.~\ref{fig:method_diagram}, we extract these features using the off-the-shelf OpenPose model \cite{cao2017realtime}. From the coordinates and confidence levels of the detected facial keypoints, our regression network estimates the apparent gaze of the corresponding subjects. From the perspective of the overall framework for ADL analysis, leveraging the facial keypoints is beneficial because a single feature extractor module can be used for two required tasks: pose estimation and gaze estimation. 
    \item the complexity of gaze estimation varies according to the scenario, such that the quality of predictions provided by a gaze regressor is expected to vary case-by-case. For this reason, \textit{our model is designed and trained to provide an estimation of its uncertainty for each prediction of gaze direction}. To that end, we leverage concepts used by Bayesian neural networks for estimation of aleatoric uncertainty.
    \item in cases such as self-occlusion, one or more facial keypoints might not be detected, and OpenPose assigns a confidence of zero to the corresponding feature. To handle the absence of detections, \textit{we introduce the concept of Confidence Gated Units (CGU) }to induce our model to disregard detections for which a zero-confidence level is provided.
\end{itemize}

\section{Related Work} \label{sec:relwork}

Ambient assisted living applications may benefit from computer vision methods in a variety of scenarios, including safety, well-being assessment, and human-machine interaction  \cite{chaaraoui2012review,leo2017computer}. Our aim is to monitor the overall health status of a patient by observing his/her behavior, or the way he/she interacts with the environment or with others. Summarized in Section \ref{sec:facility} and detailed in \cite{VISAPP2018,chessa17}, the assisted living environment where our research takes place has been used for studies on automatic assessment of mobility information and frailty \cite{martini2018data}.
Related to our system are the methods presented in \cite{cao2017realtime,bathrinarayanan2013evaluation,zouba2010activity}, which propose different smart systems designed to monitor human behavior and way of life incorporating computer vision elements. 

Estimating the relative pose of subjects is crucial to perform high level tasks such as whole body action recognition and understanding the relationship between a person and the environment. Appearance-based pose estimation systems attempt to infer the positions of the body joints of the subjects present in a scene. Traditional methods relied on models fit to each of the individual subjects found in a given image frame \cite{zhang2009efficient,brubaker2006physics}. More recent approaches employ convolutional architectures \cite{Wei2016ConvolutionalPM, cao2017realtime} to extract features from the entire scene, therefore making the whole process relatively independent of the number of subjects in the scene.



At a finer level, the analysis of human facial features may provide additional information \cite{baltruvsaitis2016openface} about well-being. For example, facial expression recognition \cite{Lopes2017Facial, Zhang2017Facial} can be used in sentiment analysis \cite{Jayalekshmi2017Facial}. Facial analysis can also provide information on gaze direction, which is useful to better understand the interaction between a person and his/her surrounding environment \cite{Varadarajan2018}. Recent contributions in this area attempt to infer the orientation of a person's head by fitting a 3D face model to estimate both 2D \cite{Zhang2015Appearance} and 3D gaze information \cite{zhang2017written}. Other contemporary methods resort to different types of information, which include head detection, head orientation estimation, or contextual information about the surrounding environment \cite{Murphy2009Headpose}. In the context of human-computer interaction, the work in \cite{kafka2016eye} employs an end-to-end architecture to track the eyes of a user in real-time using hand-held devices.

However, most works and datasets on inference of head orientation and gaze focus on specific scenarios, such as images containing close-up views of subjects' heads \cite{FunesMora_ETRA_2014,Zhang2015Appearance}, with restricted background size and complexity. More similar to our scenario of interest, the GazeFollow dataset introduced in \cite{recasens2015were} contains more than 120k images of one or more individuals performing a variety of actions in relatively unconstrained scenarios. Together with the dataset, the authors introduce a two-pathway architecture that combines contextual cues with information about the position and appearance of the head of a subject to infer his/her gaze direction. A similar model is introduced in \cite{chong2018connecting}, with applicability extended to scenarios where the subject's gaze is directed somewhere outside the image.

Gaze estimation is a task with multiple possible levels of difficulty, which vary according to the scenario of observation. Even for humans, it is much easier to tell where someone is looking if a full-view of the subject's face is possible, while the task becomes much harder when the subject is facing backwards with respect to the observer's point of view. In modeling terms, this corresponds to heteroscedastic uncertainty, i.e. uncertainty that depends on the inputs to the model, such that some inputs are associated to more noisy outputs than others. 

As explained in \cite{kendall2017uncertainties}, conventional deep learning models do not provide estimations of uncertainties for the outputs. Classification models typically employ softmax in their last layer, such that prediction scores are normalized and do not necessarily represent uncertainty. For regression models, usually no information on prediction confidence is provided by the model. Bayesian deep learning approaches are becoming increasingly more popular as a way to understand and estimate uncertainty with deep learning models \cite{gal2016uncertainty,kendall2015bayesian,kendall2018multi}. Under this paradigm, uncertainties are formalized as probability distributions over model parameters and/or outputs. For estimation of heteroscedastic uncertainty in regressors models, outputs can be modeled as corrupted with Gaussian random noise. Then, as we detail in Eq.\ref{eq:losscos} of Section \ref{sec:training}, a customized loss function is sufficient for learning a regressor model that also predicts the variance of this noise as a function of the input \cite{kendall2017uncertainties}, without need for uncertainty labels. 

\section{Proposed Approach}

Our method estimates a person's apparent gaze direction according to the relative locations of his/her facial keypoints. As Fig.~\ref{fig:method_diagram} indicates, we use OpenPose \cite{cao2017realtime} to detect the anatomical keypoints of all the persons present in the scene. Of the detected keypoints, we consider only those located in the head (i.e., the nose, eyes, and ears) of each individual.

Let $p_{k,s}^j = [x_{k,s}^j,y_{k,s}^j,c_{k,s}^j]$ represent the horizontal and vertical coordinates of a keypoint $k$ and its corresponding detection confidence value, respectively. The subscript $k\in \{n,e,a\}$ represents the nose, eyes, and ears features, with the subscript $s\in \{l,r,\emptyset \}$ encoding the side of the feature points.

Aiming at a scale-invariant representation, for each person $j$ in the scene we centralize all detected keypoints with respect to the head-centroid $h^j=[x_h^j,y_h^j]$, which is computed as the mean coordinates of all head keypoints detected in the scene. Then, the obtained relative coordinates are normalized based on the distance of the farthest keypoint to the centroid. In this way, for each detected person we form a feature vector $f\in \mathbb{R}^{15}$ by concatenating the relative vectors $\hat{p}_{k,s}^j=[\hat{x}_{k,s}^j,\hat{y}_{k,s}^j,c_{k,s}^j]$
\begin{equation}
f^j = \left[ \hat{p}_{n,\emptyset}^j,\hat{p}_{e,r}^j,\hat{p}_{e,l}^j,\hat{p}_{a,r}^j, \hat{p}_{a,l}^j \right].
\label{eq:feature}
\end{equation}

%
\subsection{Network architecture using Gated units}
Images acquired from assisted living environments can contain multiple people performing different activities, such that their apparent pose may vary significantly and self-occlusions frequently occur. For example, in lateral-views at least an ear is often occluded, while in back-views nose and eyes tend to be occluded. As consequence, an additional challenge intrinsic to this task is the representation of missing keypoints. 
In such cases, OpenPose outputs $0$ for both the spatial coordinates $(x,y)_{k,s}^j$ and also the detection confidence value $c_{k,s}^j$. Since the spatial coordinates are centralized with respect to the head-centroid $h^j$ as the $(0,0)$ reference of the input space, a confidence score $c_{k,s}^j = 0$ plays a crucial role in indicating both the reliability and also the absence of a keypoint.

\begin{figure}[h]
    \centering
    \includegraphics[width=0.6\linewidth]{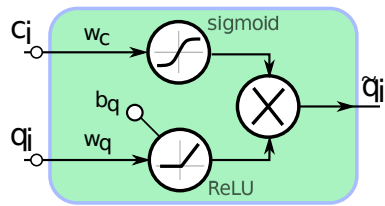}
    \caption{The proposed Confidence Gated Unit (CGU).}
    \label{fig:cgu}
\end{figure}

Inspired on the Gated Recurrent Units (GRUs) employed in recurrent neural networks \cite{cho2014learning}, we propose a Confidence Gated Unit (CGU) composed of two internal units: i) a ReLU unit acting on an input feature $q_i$; and ii) a sigmoid unit to emulate the behavior of a gate according to a confidence value $c_i$. As depicted in Figure \ref{fig:cgu}, we opt for a sigmoid unit without a bias parameter, to avoid potential biases towards models that disregard $c_i$ when trained with unbalanced datasets where the majority of samples are detected with high confidence. Finally, the outputs of both units are then multiplied into an adjusted CGU output $\tilde{q_i}$.

For our application, a CGU is applied to each pair coordinate-confidence $(\hat{x}_{k,s}^j,c_{k,s}^j)$ and $(\hat{y}_{k,s}^j,c_{k,s}^j)$.
To properly exploit the full range of the sigmoid function and thus reach output values near $0$ for $c_{k,s}^j=0$, we centralize and standardize the input confidence scores according to the corresponding dataset statistics. In this way, our proposed network for gaze regression has a combination of $10$ CGUs as input layer.

Moreover, the variety of view-points from which a subject might be visible in the scene, occlusions and unusual poses lead to a vast range of scenarios where the difficulty of the gaze estimation varies significantly. Hence, we 
design a model that incorporates an uncertainty estimation method, which indicates its level of confidence for each prediction of gaze direction. From an application perspective, this additional information would allow us to refine the predictions by choosing between different cameras, models, or time instants.

The gaze direction is approximated by the vector $\tilde{g}^j=\left[\tilde{g}_x,\tilde{g}_y \right]$, which consists of the projection onto the image plane of the unit vector centered at the centroid $h^j$. In terms of architecture design, this corresponds to an output layer with $3$ units: two that regress the $(\tilde{g}_x,\tilde{g}_y)$ vector of gaze direction, and an additional unit that outputs the regression uncertainty $\sigma_{\tilde{g}}$.

Following ablative experiments and weight visualization to identify dead units, we opt for an architecture where the CGU-based input layer is followed by $2$ fully-connected (FC) hidden layers with $10$ units each, and the output layer with $3$ units. Thus, the architecture has a total of $283$ learnable parameters and can be summarized as: (10 CGU, 10 FC, 10 FC, 3 FC).


\subsection{Training Strategy}
\label{sec:training}
While all weights composing fully-connected layers are initialized as in \cite{He2015}, we empirically observed better results when initializing the parameters composing CGU units with \textit{ones}. Since these compose only the input layer, initializing the weights as such does not represent a risk of gradient explosion as no further backpropagation has to be performed. Intuitively, our rationale is that the input coordinate features should not be strongly transformed in this first layer, as at this initial point no information from additional keypoints is accessible. Regarding regularization, we empirically observed better results without regularization in the input and output layers, while a L2 penalty of $10^{-4}$ is applied to parameters of both FC hidden layers.

Regardless of the dataset, we trained our network only on images where at least two facial keypoints are detected. Since we are interested on estimating direction of gaze to verify whether any object of interest is within a person's field of view, we opt for optimization and evaluations based on angular error. Thus, training was performed using a cosine similarity loss function that is adjusted based on \cite{kendall2017uncertainties} to allow uncertainty estimation. Let $\mathcal{T}$ be the set of annotated orientation vectors $g$, while $\tilde{g}$ corresponds to the estimated orientation produced by the network and $\sigma_{\tilde{g}}$ represents the model's uncertainty prediction. Our cost function is then given by
\begin{equation}\label{eq:losscos}
    \mathcal{L}_{\text{cos}}(g,\tilde{g}) = \frac{1}{\left|\mathcal{T}\right|}\sum_{g\in\mathcal{T}}\frac{\exp(-\sigma_{\tilde{g}})}{2} \frac{-{g}\cdot{\tilde{g}}}{||{g}||\cdot||{\tilde{g}}||} + \frac{log\:\sigma_{\tilde{g}}}{2}.
\end{equation}

With this loss function, no additional label is needed for the model to learn to predict its own uncertainty. The $\exp(-\sigma_{\tilde{g}})$ component is a more numerically stable representation of $\frac{1}{\sigma_{\tilde{g}}}$, which encourages the model to output a higher $\sigma_{\tilde{g}}$ when the cosine error is higher. On the other hand, the regularizing component $\log(\sigma_{\tilde{g}})$ helps avoiding an exploding uncertainty prediction.

In terms of model optimization, all experiments were performed using the Adam \cite{kinga2015method} optimizer with early stopping based on angular error on the corresponding validation sets. Additional parameters such as batch size and learning rate varied according to the dataset. Hence, we describe them in detail in Section \ref{sec:results}.


\section{Experiments and Results} \label{sec:results}
We evaluate our approach on two different datasets. The first is the GazeFollow dataset \cite{recasens2015were}, on which we compare our method against two different baselines. The second dataset, which we refer to as the \textit{MoDiPro} dataset, comprises images acquired from an actual discharge facility as detailed in Section \ref{sec:facility}. 

\subsection{Evaluation on the GazeFollow dataset}
{\bf Dataset split and training details.} The publicly available GazeFollow dataset contains more than 120k images, with corresponding annotations of the eye locations and the focus of attention point of specific subjects in the scene. We use the direction vectors connecting these two points to train and evaluate our regressors. In terms of angular distribution, about $53\%$ of the samples composing the GazeFollow training set correspond to subjects whose gaze direction lies within the quadrant $[-90^\circ,0^\circ]$ with respect to the horizontal axis. On the other hand, only $29\%$ of the cases their gaze direction is within the $[-180^\circ,-90^\circ]$ quadrant. To compensate such bias, we augment the number of samples in the later quadrant by mirroring with respect to the vertical-axis a subset of randomly selected samples from the most frequent quadrant. Finally, for training our model we split the training set into two subsets: $90\%$ for \textit{train}, and $10\%$ for validation \textit{val} subset. Training is performed using a learning rate $5\times10^{-3}$, batches of $1024$ samples and early-stopping based on angular error on the \textit{val} subset.
The \textit{test} set comprises $4782$ images, with ten different annotations per image. For evaluation, we follow \cite{recasens2015were} and assess each model by computing the angular error between their predictions and the average annotation vector. 

The GazeFollow dataset is structured such that for each image only the gaze from a specific subject must be assessed. For images containing multiple people, this requires identifying which detection provided by OpenPose corresponds to the subject of interest. To that end, we identify which detected subject has an estimated head-centroid that is the closest to the annotated eye-coordinates $E_{GT}$ provided as ground-truth. To avoid mismatches in cases the correct subject is not detected but detections for other subjects on the scene are available, we impose that gaze is estimated only if $E_{GT}$ falls within a radius of $1.5\times \delta$ around the head-centroid, where $\delta$ corresponds to distance between the centroid and its farthest detected facial keypoint.

We compare our method against two baselines. The first, which we refer to as \textsc{Geom}, relies solely on linear geometry to estimate gaze from the relative facial keypoints positions. Comparison against this baseline aims at evaluating if training a network is needed to approximate the regression $f \rightarrow g$, instead of directly approximating it by a set of simple equations. The second baseline is the model introduced together with the GazeFollow dataset in \cite{recasens2015were}, which consists of a deep neural-network that combines a gaze pathway and a saliency pathway that are jointly trained for gaze estimation. We refer to this baseline as \textsc{GF-model}. 


{\bf Comparison against geometry-based baseline.} We refer the reader to our Supplementary Material for a more detailed description of \textsc{Geom}. This baseline is a simplification of the model introduced in \cite{gee1994determining} for face orientation estimation, which makes minimal assumptions about the facial structure \cite{gee1994determining} but additionally requires mouth keypoints and pre-defined model ratios. In short, let $\Vec{s}$ represent the facial symmetry axis that is computed as the normal of the eye-axis. We estimate the facial normal $\Vec{n}$ as a vector that is normal to $\Vec{s}$ while intersecting $\Vec{s}$ at the detected nose position. Then, the head pitch $\omega$ is estimated as the angle between the ear-centroid and the eye-centroid, i.e., the average coordinates of eyes and ears detections, respectively. Finally, gaze direction is estimated by rotating $\Vec{n}$ with the estimated pitch $\omega$.

\begin{figure}[!h]
\centering
	\includegraphics[trim={1cm 0 0.cm 0},clip,height=2.7cm]{00004763_gaze.png}
	\hskip 2pt
	\includegraphics[trim={0 0 7.5cm 0},clip,height=2.7cm]{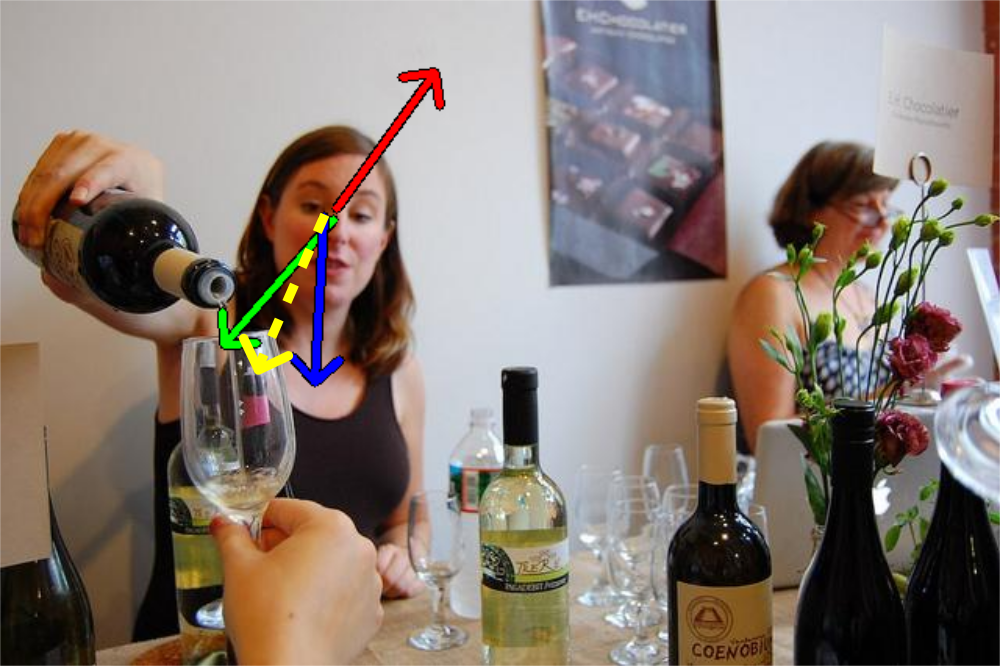}
	\hskip 2pt
	\includegraphics[trim={0 0 0 3cm},clip,height=2.7cm]{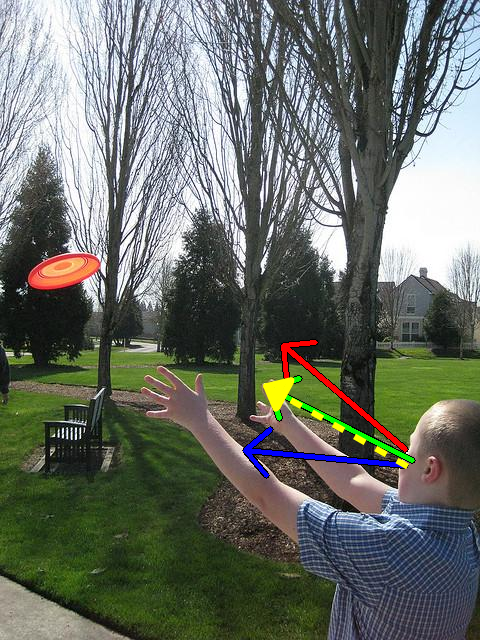}
    \begin{tikzpicture}
	\draw [line width=0.8mm, red,left] (0,0) -- (.5,0) node [right,color=black] (text1) {\footnotesize{\textsc{Geom}}};;
	\draw [line width=0.8mm, blue] (text1.east) -- ([xshift=4mm]text1.east) node [right,color=black] (text2) {\footnotesize{\textsc{GF-model}}};;
	\draw [line width=0.8mm, green] (text2.east) -- ([xshift=4mm]text2.east) node [right,color=black] (text3) {\footnotesize{\textsc{Net} (ours)}};;
	\draw [line width=0.8mm, yellow] (text3.east) -- ([xshift=4mm]text3.east) node [right,color=black] (text3) {\footnotesize{Avg. annotation}};;	
    \end{tikzpicture}       	
    \caption{Examples of gaze direction estimations provided by the different models evaluated on GazeFollow.}
	\label{fig:example_gazefollow}
\end{figure}
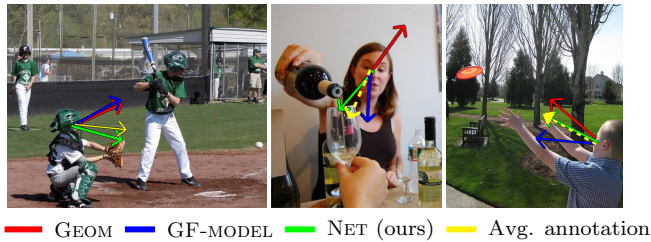

The \textsc{Geom} baseline requires the detection of the nose and at least one eye. Out of the $4782$ images composing the GazeFollow \textit{test} set, \textsc{Geom} is thus restricted to a subset $Set1$ of $4258$ images. As summarized on Tab.~\ref{tab:res_gf}, results obtained on subset $Set1$ demonstrate that our model $Net$ provide gaze estimations on average $23^\circ$ more accurate than the ones obtained with the simpler baseline. Such a large improvement in performance suggests our network learns a more complex (possibly non-linear) relationship between keypoints and gaze direction. Examples available on Fig.~\ref{fig:example_gazefollow} qualitatively illustrate how the predictions provided by our \textsc{Net} model (in green) are significantly better than the ones provided by the baseline \textsc{Geom} (in red). 
\begin{table}[!h]
\centering
    \begin{tabular}{lccc}
    \cline{2-4}
     & \multicolumn{1}{c}{\textit{{Set1}}} & \multicolumn{1}{c}{\textit{{Set2}}} & \multicolumn{1}{c}{\textit{{Full}}} \\ 
     \textit{No. of images} & 4258 & 4671 & 4782
     \\\hline \hline
    \textsc{Geom} & $42.63^\circ$ & - & - \\
    \textsc{Net0} & $19.52^\circ$ & $25.70^\circ$ & -  \\
    \textsc{Net} & $19.41^\circ$ & $23.37^\circ$ & -  \\
    \textsc{GF-model\cite{recasens2015were}} & - & - & $24^\circ$ \\ \hline
    \end{tabular}
\caption{Comparison in terms of angular errors between our method and baselines on the GazeFollow test set.}
\label{tab:res_gf} 
\end{table}

\begin{figure*}[t]
    \centering
    \includegraphics[width=\linewidth]{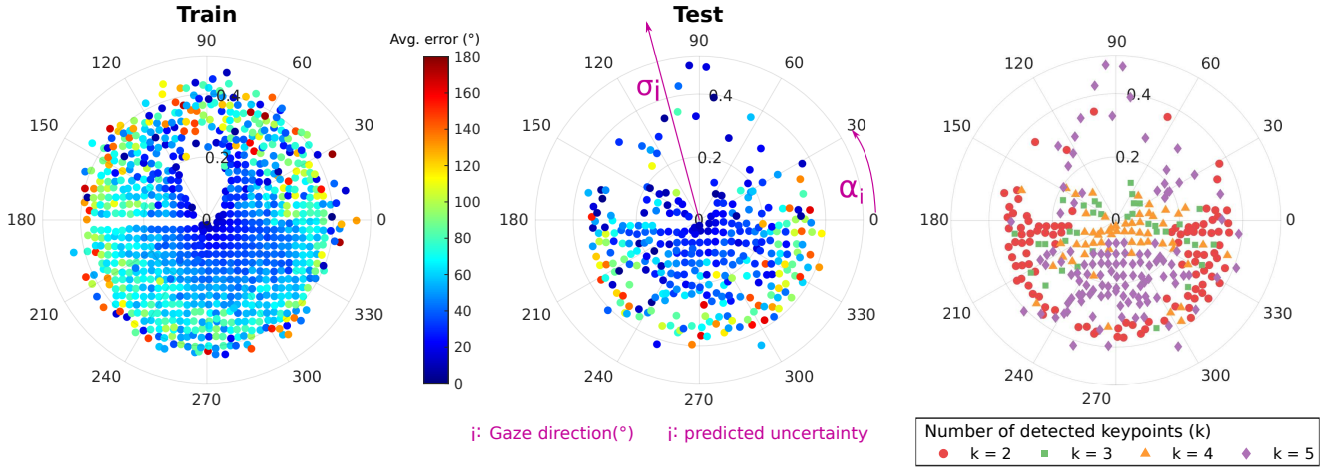}
    \caption{Distribution of gaze direction ($\alpha_i$) and uncertainty predictions ($\sigma_i$) provided by our proposed model. \textit{Left/center}: colormap depicts angular error of predictions. \textit{Right}: colors represent the amount of keypoints detected by OpenPose for corresponding samples. For better visualization, samples are grouped into equally spaced bins.}
    \label{fig:polarplots}
\end{figure*}
{\bf Comparison against GazeFollow model.} Since our network is trained on images where at least two facial keypoints are detected, we apply the same constraint for evaluation. In the test set, OpenPose detects at least two keypoints for a subset $Set2$ containing $97.7\%$ of the $4782$ images composing the full set. 

The results of our evaluation are summarized in Tab.~\ref{tab:res_gf}, while qualitative examples are provided in Fig.~\ref{fig:example_gazefollow}. As reported in \cite{recasens2015were}, gaze predictions provided by the \textsc{GF-model} present an mean angular error of $24^\circ$ on the \textit{test} set. Our \textsc{Net} model provides an mean angular error of $23.37^\circ$ for $97.7\%$ of these images, which strongly indicates that its performance is on pair with \textsc{GF-model} network despite relying solely on the relative position of $5$ facial keypoints to predict gaze. 

{\bf Impact of using Confidence Gated Units (CGU).} To verify the benefits of applying our proposed CGU blocks to handle absent keypoint detections, i.e. keypoints with $0$ confidence score, we evaluated the performance of our model with and without feeding the confidence scores as inputs. We refer to the latter case as the \textsc{Net0}, where the CGU blocks composing the input layer are replaced by simple ReLU units initialized in the same way as described in Section \ref{sec:training}. Results summarized in Tab.~\ref{tab:res_gf} indicate an error decrease of $2.3^\circ$ when providing confidence scores to an input layer composed of CGUs. In addition to experiments summarized in Tab.~\ref{tab:res_gf}, we also evaluated a model where the CGU units are replaced by simple additional ReLU units to handle confidence scores. For the $1536$ images where OpenPose detects less than $4$ facial keypoints, a significant decrease on angular error is observed when using CGU units: $30.1^\circ$ mean error, in comparison to $30.9^\circ$ provided by the model with solely ReLU based input layer.

\begin{figure}[!h]
    \centering
    \includegraphics[trim={0.2cm 0 0.2cm 0},clip,width=\linewidth]{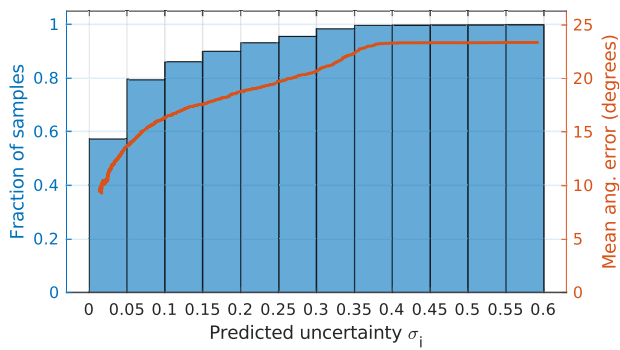}
    \caption{Cumulative mean angular error according to uncertainty predicted by our model for each sample. }
    \label{fig:cum_unc}
\end{figure}

{\bf Quality of uncertainty estimations.} In addition to the overall mean angular error, we also evaluate how accurate are the uncertainty estimations provided by our \textsc{Net} model for its gaze direction predictions. As depicted in Fig.~\ref{fig:cum_unc}, significantly lower angular errors are observed for gaze predictions accompanied by low uncertainty network predictions. Uncertainties lower than $0.1$ are observed for $80\%$ of the \textit{test} set, a subset for which the gaze estimations provided by our \textsc{Net} model are on average off by only $16.5^\circ$.

Moreover, the high correlation between uncertainty predictions and angular error ($\rho=0.56$) is clearly depicted by the plots provided in Fig.~\ref{fig:polarplots}. For each sample in these plots, the radial distance corresponds to its predicted uncertainty $\sigma_i$, while the angle corresponds to predicted direction of gaze $\tilde{g}$, i.e $\alpha_i=tan^{-1}(-\tilde{g}_y/\tilde{g}_x)$. For both \textit{train} and \textit{test} sets, the associated colormap shows that lower errors (in dark blue) are observed for predictions with lower uncertainty, with increasingly higher errors (green to red) as the uncertainty increases (farther from the center). 

{\bf Performance according to keypoint occlusions.} Furthermore, the central and the right-most scatter plots in Fig.~\ref{fig:polarplots} also allow an analysis on how the performance of our model and its uncertainty predictions vary according to specific scenarios. For most cases, the number of detected keypoints ($k$) indicates specific scenarios: $k=2$ is mostly related to back-views, where nose and two other keypoints (both eyes or a pair eye-ear) are missing; $k=3$ and $k=4$ are mostly lateral-views; $k=5$ are frontal-views, where all keypoints are visible. Since images are $2D$ projections from the environment, back- and frontal-views are the ones more affected by the information loss implicit in the image formation process, while for lateral-views estimation of gaze direction tends to be easier. 

An analysis of the scatter plots demonstrates that the predictions provided by our model reflect these expected behaviors. For samples with $k=2$ (back-view), both uncertainty predictions and angular error tend to be higher, while for most cases of $k=3$ and $k=4$ the predictions are associated with lower uncertainty and higher angular accuracy. Moreover, the spread of the distribution of predictions for $k=5$ indicates that the model's uncertainty predictions are not just correlated to the amount of available keypoints and predicted gaze direction, but rather case specific while still highly correlated to angular errors. 


\subsection{Results on the assisted living dataset}
\label{sec:facility}

{\bf Dataset split and training details.} This work is part of a project that focuses on elderly patients with partial autonomy but in need of moderate assistance, possibly in a post-hospitalization stage. Thus, it is critical to evaluate the performance of our gaze estimation model on data from real assisted living environments. To that end, we also evaluate our approach on videos acquired in an assisted living facility in which the patient, after being discharged from the hospital, is hosted for a few days. The facility is a fully-equipped apartment situated in a local hospital
, where patients may be monitored by various sensors, including localization systems, RGB-D, and two conventional video cameras, arranged as shown in Fig.~\ref{fig:apt_plan}. 


More specifically, to evaluate the performance of our gaze estimation model we compiled a dataset which we call \textit{MoDiPro}, consisting of 1,060 video frames collected from the two cameras whose positions are indicated in Fig.~\ref{fig:apt_plan}. For $CAM1$, $530$ frames were sampled from $46$ different video sequences; for $CAM2$, $530$ frames were sampled from $27$ different video sequences. To limit storage while discarding minimal temporal information, the resolution of the acquired frames was limited to $480\times 270$ pixels, at $25$ fps. In most frames multiple subjects are simultaneously visible, with a total of 22 subjects performing different activities. 

\begin{figure}[t]
	\centering
\begin{tikzpicture}
   \node[inner sep = 0pt] (a) {\includegraphics[trim={0 1.2cm 5cm 0}, clip, width=0.49\linewidth]{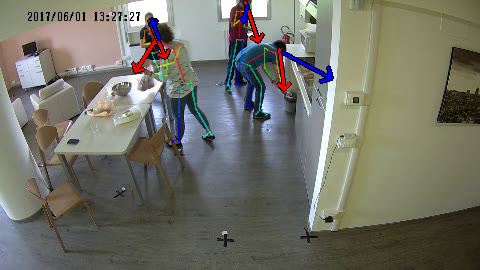}};
    \node[fill,above left,inner sep = 0pt] at (a.south east) {\colorbox{black}{\textcolor{white}{\large CAM1}}};
\end{tikzpicture} \hfill 
\begin{tikzpicture}
   \node[inner sep = 0pt] (a) {\includegraphics[trim={2cm 1.5cm 3.5cm 0}, clip, width=0.49\linewidth]{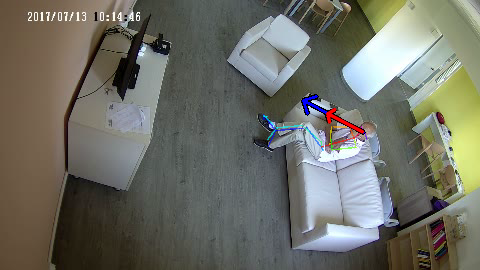}};
    \node[fill,below left,inner sep = 0pt] at (a.north east) {\colorbox{black}{\textcolor{white}{\large CAM2}}};
\end{tikzpicture} \hfill \\ 
\vskip 0.5em
\begin{tikzpicture}
   \node[inner sep = 0pt] (a) {\includegraphics[trim={0 1.2cm 5cm 0}, clip, width=0.49\linewidth]{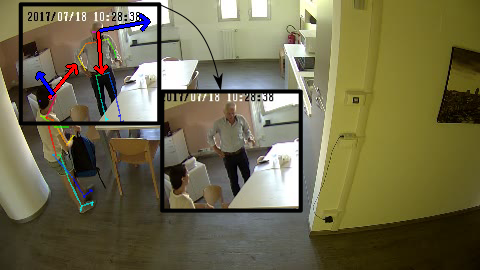}};
    \node[fill,below left,inner sep = 0pt] at (a.north east) {\colorbox{black}{\textcolor{white}{\large CAM1}}};
\end{tikzpicture} \hfill
\begin{tikzpicture}
   \node[inner sep = 0pt] (a) {\includegraphics[trim={1.5cm 3.5cm 7cm 0}, clip, width=0.49\linewidth]{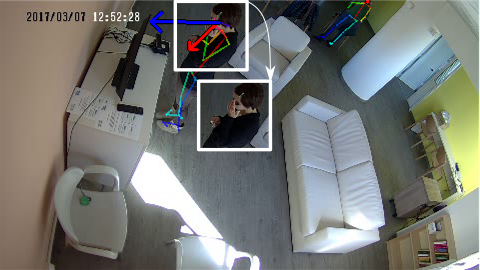}};
    \node[fill,above right,inner sep = 0pt] at (a.south west) {\colorbox{black}{\textcolor{white}{\large CAM2}}};
\end{tikzpicture} 
    \begin{tikzpicture}
    	\draw [line width=0.8mm, red,left] (0,0) -- (.5,0) node [right,color=black] (text1) {\footnotesize{\textsc{Net} (ours)}};;
    	\draw [line width=0.8mm, blue] (text1.east) -- ([xshift=4mm]text1.east) node [right,color=black] (text2) {\footnotesize{\textsc{GF-model}}};;
    \end{tikzpicture}  \hfill
	\caption{Examples of results for our gaze direction estimation approach in the \textit{MoDiPro} dataset.}
	\label{fig:example_gaze}
\end{figure}
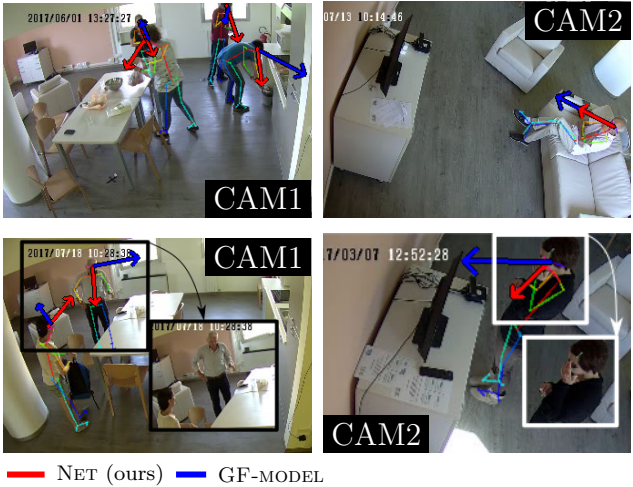

As exemplified also in Fig.~\ref{fig:example_gaze}, cameras $CAM1$ and $CAM2$ cover different parts of the environment. Images acquired with $CAM2$ present significant distortion, which increases the complexity of the task. We randomly split the available sets of images into camera-specific training, validation and test subsets. Since frames composing the same video sequence can be highly correlated, we opt for a stratified strategy where video sequences are sampled. That is, all frames available from a certain video sequence are assigned to either \textit{train}, \textit{val} or \textit{test} subsets. Aiming at an evaluation that covers a wide variety of scenes, the proportions chosen in terms of total number of frames are: $50\%$ for training, $20\%$ for validation,  $30\%$ for testing. Fine-tuning experiments are performed using learning rates $1\times10^{-5}$, while $1\times10^{-4}$ is adopted when training models only on \textit{MoDiPro} images. Batches with $64$ samples are used, with early-stopping based on angular error on the \textit{val} subset. Moreover, all results reported on Tab.~\ref{tab:res_modipro} and discussed below correspond to average values obtained after train/test on $3$ different random splits.

To assess the cross-view performance of our method, we train our \textsc{Net} model with $7$ different combinations of images from \textit{MoDiPro} and GazeFollow datasets. As summarized in Tab.~\ref{tab:res_modipro}, models \textsc{Net}\#0-2 are trained in $CAM1$-only, $CAM2$-only, and both \textit{MoDiPro} cameras. \textsc{Net}\#3 corresponds to the model trained only on GazeFollow frames (GF for shortness), while \textsc{Net}\#4-6 are obtained by fine-tuning the pre-trained \textsc{Net}\#3 on three possible sets of \textit{MoDiPro} frames.

\begin{table}[h]
    \centering
    \setlength\tabcolsep{2.5pt}%
    \begin{tabular}{llcccccc}
        \cline{2-4}\cline{5-7}
        & \multicolumn{3}{c}{\textsc{Train}} & \multicolumn{3}{c}{\textsc{Test}} \\ \cline{2-4} \cline{5-7}
        Model & GF & \textit{Cam1} & \textit{Cam2} & \textit{Cam1} & \textit{Cam2} & \textit{Mean} \\ \hline \hline
        \textsc{Net}\#0 &  &\checkmark &   & $16.16^\circ$ & $39.12^\circ$ & - \\
        \textsc{Net}\#1& & & \checkmark& $29.56^\circ$ & $26.37^\circ$ & - \\
        \textsc{Net}\#2& & \checkmark& \checkmark& $18.52^\circ$ & $23.02^\circ$ & $20.94^\circ$  \\
        \textsc{Net}\#3& \checkmark & & & $27.64^\circ$ & $26.98^\circ$ & $27.31^\circ$\\ 
        \textsc{Net}\#4& \checkmark & \checkmark& & $16.17^\circ$ & $27.36^\circ$ & - \\
        \textsc{Net}\#5& \checkmark & & \checkmark & $27.56^\circ$ & $24.01^\circ$ & - \\
        \textsc{Net}\#6& \checkmark &\checkmark &\checkmark & $17.82^\circ$ & $20.15^\circ$ & $19.05^\circ$  \\ \cline{1-7}
        \textsc{GF-model} & \checkmark & & & $43.49^\circ$ & $60.82^\circ$ & $52.15^\circ$ \\ \cline{1-7}\vspace{-5pt}
    \end{tabular}
    \caption{Performance of our method on the \textit{MoDiPro} dataset for different combinations of training/testing sets.} 
    \label{tab:res_modipro}
\end{table}

{\bf Performance according to training/testing subsets.} Cross-view results obtained by \textsc{Net}\#0 on $CAM2$ and \textsc{Net}\#1 on $CAM1$ demonstrate how models trained only on a camera-specific set of images are less robust to image distortions, with significantly higher angular errors for images composing unseen subsets. Trained on both $CAM1$ and $CAM2$, the model \textsc{Net}\#2 demonstrates a more consistent performance across views. In comparison with the camera specific models, a $3^\circ$ lower angular error on $CAM2$ is obtained at cost of only $1.4^\circ$ error increase on $CAM1$.

In addition, error comparisons between models \textsc{Net}\#0-2 and \textsc{Net}\#4-6 demonstrate that pre-training the model on the GF dataset before fine-tuning on \textit{MoDiPro} images leads to consistently lower mean angular errors, with an optimal performance of $17.82^\circ$ for $CAM1$ and $20.15^\circ$ for $CAM2$. This corresponds to an overall average error $1.9^\circ$ lower than the model \textsc{Net}\#2 not pre-trained on GF, while more than $7^\circ$ better than the model \textsc{Net}\#3 trained solely on GF. In terms of camera-specific performance, for $CAM1$ optimal performances with error below $17^\circ$ are obtained when not training on $CAM2$. On the other hand, predictions for $CAM2$ are significantly better when training is performed using additional $CAM1$ and/or GazeFollow images. We hypothesize the distortions characteristic of $CAM2$ images easily lead to overfitting, thus the advantage of training on additional sets of images. As a final remark we may notice that overall \textsc{Net}\#6 provides the best an most stable result across the two views.

{\bf Comparison against }\textbf{\textsc{GF-model}}. Finally, we compare the predictions provided by our \textsc{Net} models to the ones obtained by the publicly available version of \textsc{GF-model}\footnote{This version provides $25.8^\circ$ mean angular error on the GazeFollow test set, in comparison to the $24^\circ$ reported in \cite{recasens2015were}}. As summarized in Tab.~\ref{tab:res_modipro}, gaze predictions provided by \textsc{GF-model} on the \textit{MoDiPro} dataset are remarkably worse in terms of angular error than the ones predicted by any of our \textsc{Net}\#0-6 models, including the \textsc{Net}\#3 also trained only on GF images. 

Closer inspection of \textsc{GF-model} predictions suggests two disadvantages of this model with respect to ours when predicting gaze on images from real assisted living environments: i) sensitivity to scale; ii) bias towards salient objects. Images composing the GazeFollow typically contain a close-view of the subject of interest, such that only a small surrounding area is covered by the camera-view. In contrast, images from assisted living facilities such as the ones in the \textit{MoDiPro} dataset contain subjects covering a much smaller region of the scene, i.e., they are smaller in terms of pixel area. Our \textsc{Net} model profits from the adopted representation of keypoints, with coordinates centered at the head-centroid and normalized based on the largest distance between centroid and detected keypoints. Moreover, visual inspection of \textsc{GF-model} predictions reveals examples such as the two bottom ones in Fig.~\ref{fig:example_gaze}: in the left, while our model correctly indicates that the subjects look at each other, \textsc{GF-model} is misled by the saliency of the TV and possibly the window; in the right, the saliency of the TV again misguides \textsc{GF-model}, while our model properly indicates that the person is looking at the object she is holding.

\subsection{Runtime Analysis}
Each call to our network requires approximately 0.85 ms on average on a NVIDIA GeForce 970M, with one feedforward execution per person. The overall runtime is thus domined by OpenPose, whose overall runtime is reported in \cite{cao2017realtime} as taking $77ms$ on COCO images with a NVIDIA GeForce GTX-1080 Ti GPU.

\section{Conclusion}
This paper presents an alternative gaze estimation method that exploits solely facial keypoints detected by a pose estimation model. As our end goal is to study the behaviors of individuals in a assisted living environment, exploring a single feature extraction backbone for both pose as well as gaze estimation facilitates the design of a significantly less complex overall model.

Results obtained on the GazeFollow dataset demonstrate that our method estimates gaze with accuracy comparable to a complex task-specific baseline, without relying on any image features except relative position of facial keypoints. In contrast to conventional regression methods, our proposed model also provides estimations of uncertainty of its own predictions, with results demonstrating that the predicted uncertainties are highly correlated to the actual angular error of the corresponding gaze predictions. Moreover, analysis of performance according to the number of detected keypoints indicates that the proposed Confidence Gate Units improve the performance of our model for cases of partial absence of features.

Finally, evaluation on frames collected from a real assisted living facility demonstrate the higher suitability of our method for ADL analysis in realistic scenarios, where images cover wider areas and subjects are visible at different scales and poses. In the future, we intend to explore tracking mechanisms to estimate gaze direction in videos without abrupt estimation changes. To identify human-human and human-object interactions, we plan to combine gaze estimations with a semantic segmentation model to be designed specifically for objects of interest for ADL analysis. 

\bibliographystyle{ieee}
\bibliography{gaze_estimation}

\end{document}